\pdfoutput=1

\documentclass[11pt]{article}

\usepackage[preprint]{acl}

\usepackage{times}
\usepackage{latexsym}
\usepackage{amsmath, amssymb}

\usepackage[T1]{fontenc}

\usepackage[utf8]{inputenc}

\usepackage{microtype}

\usepackage{inconsolata}

\usepackage{graphicx}

\title{Adaptive Group Policy Optimization: Towards Stable Training and Token-Efficient Reasoning}

\author{
    Chen Li, Nazhou Liu, Kai Yang \\
    HPC-AI Tech \\
    \texttt{\{lichen,liunazhou,yangkai\}@luchentech.com}
}

\begin{document}
\maketitle

\begin{abstract}
Since DeepSeek-R1 popularized, Group Relative Policy Optimization (GRPO) has become the core part of training Reasoning LLMs. However, we find some deficiency that influences RL stability and inference efficiency, like zero-variance in advantage estimation. Thus, we propose Adaptive Group Policy Optimization (AGPO) which uses a simple but effective method, an adaptive loss function, to mitigate training fluctuation and token inefficiency. The experiments demonstrate our method achieves more stable training and superior performance with significantly fewer tokens in reasoning steps.
\end{abstract}

\section{Introduction}
Large Language Models (LLMs)\cite{bommasani2021opportunities, wei2022emergent, zhao2023survey} have achieved impressive performance through extensive pre-training and post-training processes. However, effectively generating desired model responses often necessitates aligning outputs with specific downstream tasks and human preferences\cite{wang2023aligning, wolf2023fundamental}.

For alignment challenges, reinforcement learning from human feedback (RLHF)\cite{kaufmann2023survey} is introduced as a prominent post-training strategy, adopted by notable LLMs including GPT-4, Claude, Gemini, and DeepSeek. They have explored various optimization techniques such as Proximal Policy Optimization (PPO) \cite{schulman2017proximal} and Direct Preference Optimization (DPO) \cite{rafailov2023direct}. Recently, to significantly reduce computational and memory overhead associated with PPO, DeepSeek eliminated the value model and proposed Group Relative Policy Optimization (GRPO)\cite{guo2025deepseek}, which achieved high computational efficiency and excellent reasoning performance, surpassing other open-source models ranging from 7B to 70B.

Despite the demonstrated success of GRPO, it introduces challenges that can affect stable training and inference efficiency.

\textbf{Confusing Training Signal:} When all rewards within a group are identical, the normalized advantage approaches 0, causing the loss signal to vanish, which potentially stalling training. Besides, negative losses happen in RL training, but in this scenario it is not always beneficial.

\textbf{Inefficient CoT Length:} Since GRPO lacks mechanisms to discourage excessively long chain-of-thought (CoT), models tend to produce overly verbose explanations. A refined approach that rewards concise and effective reasoning is essential to improve token efficiency.

To address these issues, we propose an enhanced training algorithm, Adaptive Group Policy Optimization (AGPO). Our main contributions are summarized as follows:

\begin{itemize}
    \item \textbf{Training Stability:} By identifying the limitations of GRPO's advantage, we introduce an adaptive loss function that addresses the scenarios of zero advantage and negative losses, ensuring continuous and effective learning.
    \item \textbf{Token Efficiency:} Our adaptive loss implicitly improves token efficiency. Compared with GRPO baselines, our approach achieves better performance with significantly fewer response tokens.
\end{itemize}

\section{Background}

\subsection{Policy Gradient}
Policy gradient method is one of the most fundamental RL algorithm that directly model and optimize the policy. For any differentiable policy, the policy gradient is :

\begin{equation}
\nabla_{\theta} J(\theta) = \mathbb
{E}_{\pi_{\theta}} 
\left[ \sum_{t=0}^{T} \nabla_{\theta} \log \pi_{\theta} (a_t | s_t) A_t \right]
\end{equation}

Where advantage \(A_t\) is the most crucial part for policy gradient method, which defines how much better a specific action \(a_t\) is, compared to average action given a state \(s_t\).

\subsection{Proximal Policy Optimization (PPO)}
PPO \citep{schulman2017proximal} is one of the policy gradient methods which uses clipped surrogate objective for policy optimization. Specifically, it maximize the following objective:

\begin{equation}
\small
J_{\text{PPO}}(\theta) = \mathbb{E}_{\pi_{\theta_{\text{old}}}} 
\left[
\min \left( r_t {A}_t, \operatorname{clip}(r_t, 1 - \epsilon, 1 + \epsilon) {A}_t \right)
\right] 
\end{equation}

Where $\epsilon$ is a hyper-parameter used for tuning clipping range. \(A_t\) is the advantage, which generally will be computed by utilizing Generalized Advantage Estimation (GAE) \citep{2015arXiv150602438S} in PPO. \(r_t\) is the probability ratio of predicting token \(o_t\) for a given question \(q\) before and after the policy update:

\begin{equation}
    r_t(\theta) = \frac{\pi_{\theta} (o_t | q, o_{<t})}{\pi_{\theta_{\text{old}}} (o_t | q, o_{<t})}
\end{equation}

\subsection{Group Relative Policy Optimization (GRPO)}
Compared to PPO, GRPO \citep{2024arXiv240203300S} significantly saves the training cost through eliminating the critic model in PPO. This is achieved by approximating the advantage \(A_i\) as group-normalized reward:

\begin{equation}
    A_i = \frac{r_i - \text{mean}(\{r_1, r_2, \dots, r_G\})}{\text{std}(\{r_1, r_2, \dots, r_G\})}
\end{equation}

Where \(\text{mean}(\{r_1, r_2, \dots, r_G\})\) and \(\text{std}(\{r_1, r_2, \dots, r_G\})\) denotes the within-group mean and standard deviation respectively.

With the estimated advantage and a KL divergence penalty term, GRPO generates a group of outputs \(\{o_i\}_{i=1}^G \) based on \(\pi_{\theta_{\text{old}}}\) for each question \(q\) and update \(\pi_{\theta}\) with following objective:

\begin{equation}
\resizebox{0.47\textwidth}{!}{$
\begin{aligned}
J_{\text{GRPO}}(\theta) &= \mathbb{E}_{(q) \sim P(Q), \{o_i\}_{i=1}^G \sim \pi_{\theta_{\text{old}}}(O | q)} \Bigg[ \frac{1}{G} \sum_{i=1}^{G} \Bigg( \min \Bigg( \frac{\pi_{\theta}(o_{i} | q)}{\pi_{\theta_{\text{old}}}(o_{i} | q)} {A}_{i},  \\
& \quad \quad \quad \operatorname{clip}\left( \frac{\pi_{\theta}(o_{i} | q)}{\pi_{\theta_{\text{old}}}(o_{i} | q)}, 1 - \epsilon, 1 + \epsilon \right) {A}_{i} \Bigg)  - \beta D_{\text{KL}}(\pi_{\theta} \parallel \pi_{\text{ref}}) \Bigg) \Bigg]
\end{aligned}
$}
\label{eq:grpo}
\end{equation}

Where $\epsilon$ and $\beta$ are hyper-parameters.

\section{Adaptive Group Policy Optimization}
Performance fluctuation and entropy collapse always happen in the later stage of GRPO training, and we want to tackle it from the perspective of loss function. The loss function of GRPO has two issues: 

\begin{itemize}
    \item \textbf{Zero advantage and thus zero loss:} It is caused by groups with all-correct or all-wrong responses. This is not effectively being ignored during optimization by GRPO. According to \autoref{eq:grpo}, the mean loss of a batch is used for model update, and zero losses in the batch lead to misestimation of the scale.
    \item \textbf{Positive advantage and thus negative loss:} We observed entropy collapse same as other works like DAPO \citep{yu2025dapo}. The entropy drops quickly leading the policy model to give up exploring. However, we consider it as the problem of negative loss rather than clip ratio. It makes the model overconfident and tends to ignore incorrect samples in the batch.
\end{itemize}

Therefore, we propose an \textbf{adaptive loss} to tackle these issues as shown in \autoref{eq:objective}. By replacing the original objective, our method focuses on useful information in the batch and normalizes the loss for stable training, which brings performance improvement and token efficiency.

\begin{equation}
    \resizebox{0.45\textwidth}{!}{$
        L_{(q) \sim P(Q)} =
        \left\{
        \begin{array}{ll}
        masked, & \text{if } \{o_i\}_{i=1}^G \text{ all correct or wrong} \\
        max(0, -J_{\text{GRPO}}(\theta)), & \text{otherwise}
        \end{array}
        \right.
    $}
\label{eq:objective}
\end{equation}

\subsection{Loss Mask}
It is easy to find that if the response rewards are all equal in a group, like all correct or all wrong, then the advantages become all zero. The phenomenon of zero advantages naturally becomes more often during training as the model becomes better. Usually in RL training, the model is updated through a mini batch of training data. The corresponding groups in the batch have no effect in the training. Loss mask does not aim to reduce the percentage of such groups, but to avoid including them in the loss computation. Therefore, the loss mask in the first part of \autoref{eq:objective} means excluding invalid samples in the mean operation of sample losses.

\subsection{Loss Clip}
Clipping negative batch loss to zero is a simple but effective way to penalize the overconfidence, which achieves solid performance improvements. Another work \citep{zhu2025surprising} has similar conclusion that training with only negative samples can be highly effective.

\section{Experiments}
We conduct a few experiments for evaluating how our method affects the RL training of reasoning models.

\subsection{Implementation Details}
We use Qwen2.5-7B and Qwen2.5-14B as the base models. All experiments are conducted on our curated dataset, which is constructed by mixing data from MATH train set \citep{hendrycks2021measuring} and DAPO train set \citep{yu2025dapo}. We remove overly easy and overly hard data points from these sets. Specifically, for MATH train set, only data where difficulty levels are greater than 2 are selected. Similarly, for DAPO train set, data are retained only if solution rates achieved by Qwen3-32B model \citep{yang2025qwen3} fell between 0.5 and 0.8 inclusively.

VeRL \citep{sheng2024hybridflow} is utilized to perform RL training with a train batch size of 32, a PPO mini batch size of 8 and a learning rate of \(1e-6\). The number of group rollout is 8. Temperature for generation is set to 1. As for reward settings, the accuracy rewards \(r_{correct}\) and \(r_{wrong}\) are set to 0 and 1 respectively. Checkpoints that achieve best performance on MATH-500 \citep{lightman2023let} with the metric of Pass@1 are selected for further evaluation with respect to token efficiency.

It is worth noting that KL divergence penalty is not applied for all experiments. This is based on the observation that model distribution can vary significantly compared to reference model during long CoT training. Therefore, removing KL divergence has been adopted as common practice in the domain \citep{schulman2017proximal, yu2025dapo}.

\subsection{Main Results}
\label{sec:training}

\begin{table}[]
\resizebox{\columnwidth}{!}{
\begin{tabular}{lc}
\hline
\textbf{Model}                            & \textbf{MATH-500 (Pass@1)} \\ \hline
Qwen2.5-7B                                & 44.0                       \\
Qwen2.5-7B-GRPO                           & 73.2                       \\
Qwen2.5-7B-AGPO                           & 74.6                       \\
Qwen2.5-7B-AGPO (w/o loss mask)           & 73.4                       \\
Qwen2.5-7B-AGPO (w/o loss clip)           & 73.0                       
\\ \hline
Qwen2.5-14B                               & 59.8                       \\
Qwen2.5-14B-GRPO                          & 75.4                       \\
Qwen2.5-14B-AGPO                          & 77.2                       \\
Qwen2.5-14B-AGPO (w/o loss mask)          & 75.0                       \\
Qwen2.5-14B-AGPO (w/o loss clip)          & 75.2
\\ \hline
\end{tabular}
}
\caption{Performance of different RL techniques on MATH-500.}
\label{tab:performance}
\end{table}

\begin{table}[]
\resizebox{\columnwidth}{!}{
\begin{tabular}{lc}
\hline
\textbf{Model}           & \textbf{Average Response Tokens} \\ \hline
Qwen2.5-7B               & 571                              \\
Qwen2.5-7B-GRPO          & 699                              \\
Qwen2.5-7B-AGPO          & 533
\\ \hline
Qwen2.5-14B              & 772                              \\
Qwen2.5-14B-GRPO         & 574                              \\
Qwen2.5-14B-AGPO         & 521                                  \\ \hline
\end{tabular}
}
\caption{Average response length of different RL techniques on MATH-500.}
\label{tab:token}
\end{table}

\autoref{tab:performance} shows the performance of different models on the benchmark. Both GRPO and AGPO acquire huge performance gains compared with the base models. As for Qwen2.5-7B experiments, we observe a clear improvement on MATH-500 from 73.2 to 74.6 at the best checkpoint. Qwen2.5-14B experiments also give same conclusion that AGPO further refines the model by 1.8 percentage. It is obvious that both loss mask and loss clip are important for our method. If we train without the mask, the performance drops even lower than that of GRPO for all model sizes and it is same for loss clip.

\autoref{tab:token} illustrates the token efficiency of different models on the benchmark. Qwen2.5-7B-AGPO only takes 533 tokens on average for solving MATH-500 problems while Qwen2.5-7B-GRPO consumes 699 tokens that is \(31\%\) higher. In terms of 14B models, our AGPO also achieves best in token efficiency which uses 521 tokens averagely.

\subsection{Training Dynamics}

\begin{figure}[htbp]
    \centering
    \includegraphics[width=0.5\textwidth]{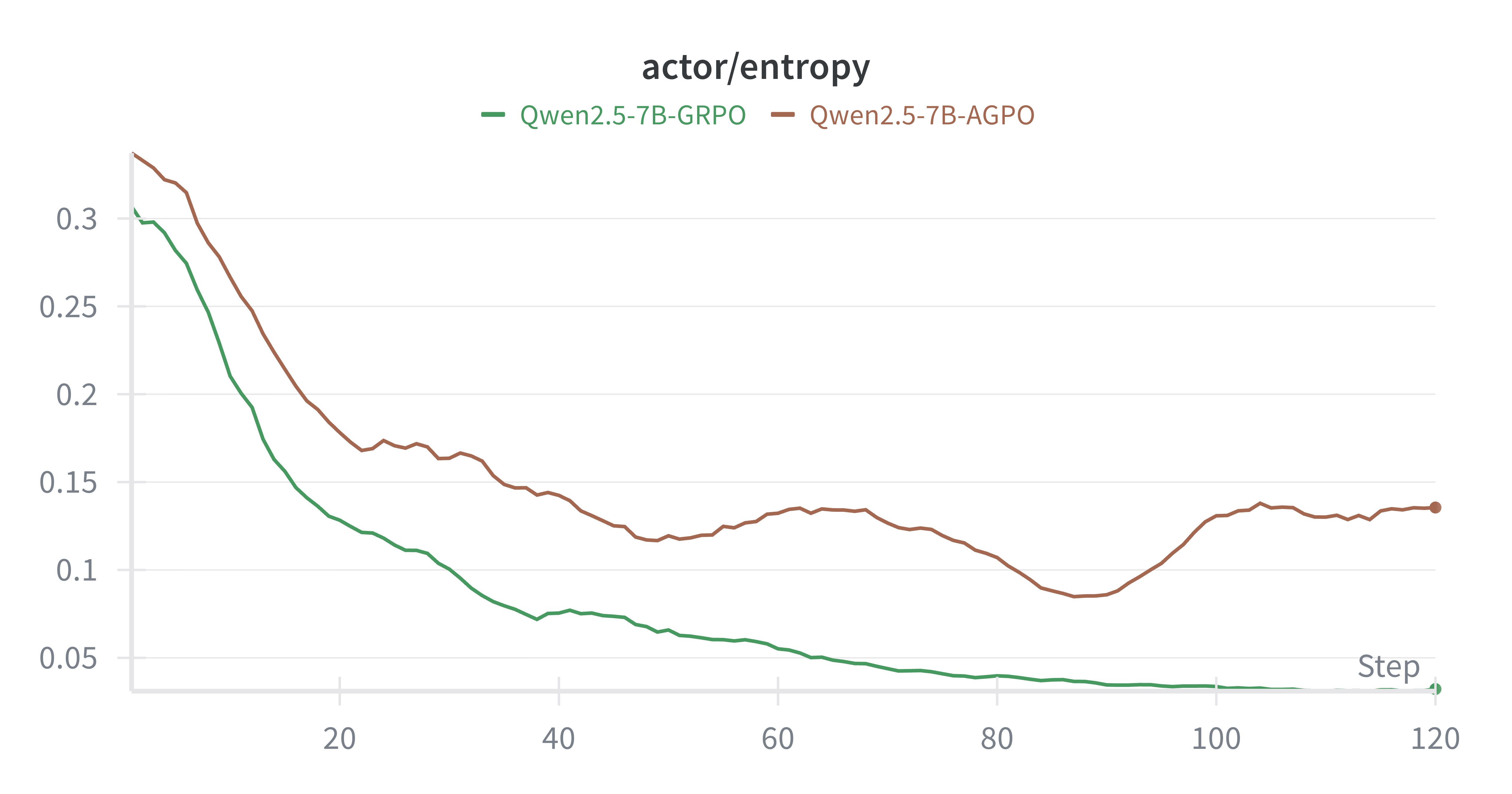}
    \caption{Actor entropy curves of GRPO and AGPO for Qwen-2.5-7B}
    \label{fig:7B_entropy}
\end{figure}

\begin{figure}[htbp]
    \centering
    \includegraphics[width=0.5\textwidth]{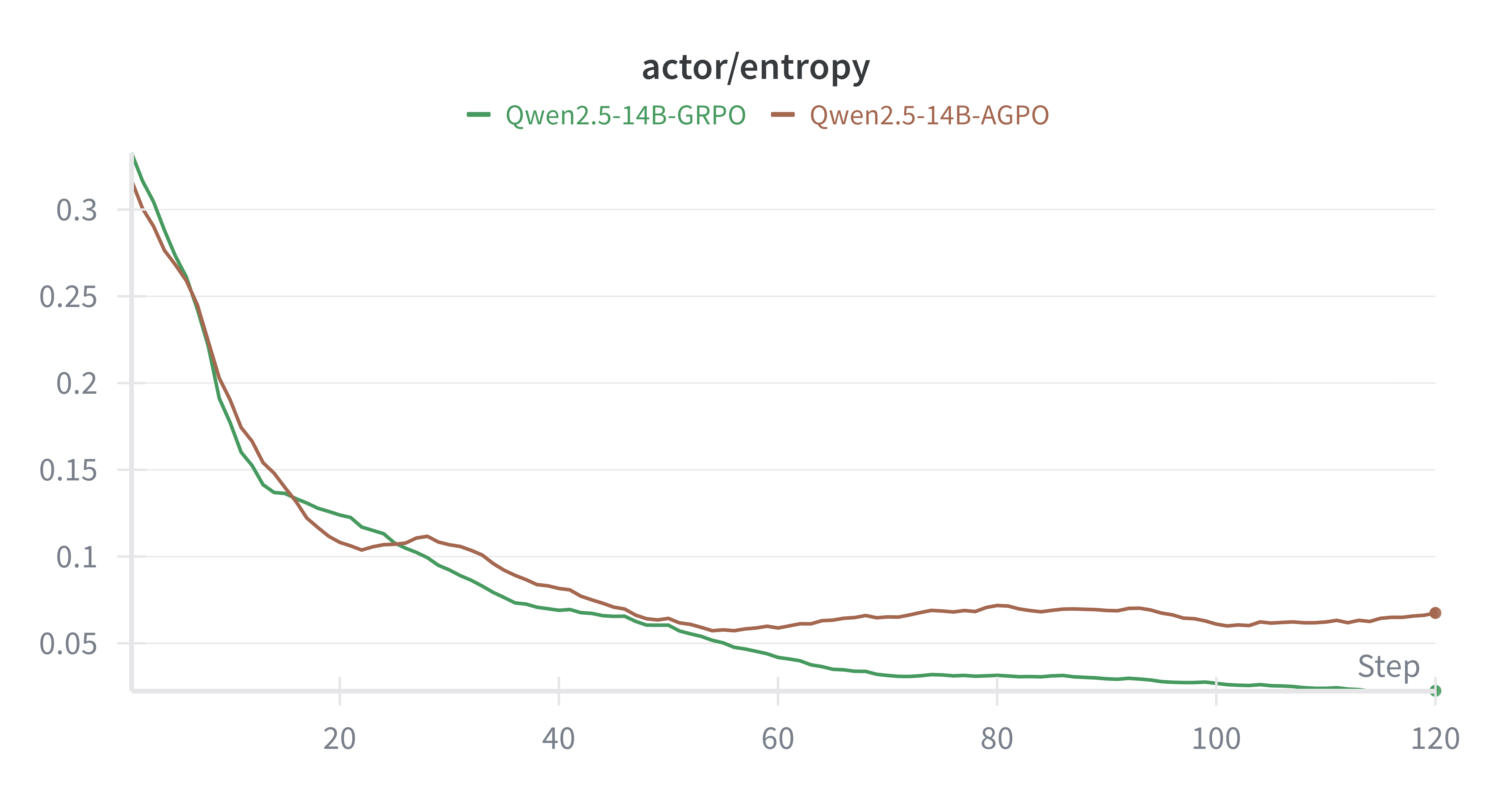}
    \caption{Actor entropy curves of GRPO and AGPO for Qwen-2.5-14B}
    \label{fig:14B_entropy}
\end{figure}

We examine several training metrics of AGPO for further analysis.

As shown in \autoref{fig:7B_entropy} and \autoref{fig:14B_entropy}, a significant enhancement to entropy is observed for both Qwen2.5-7B and Qwen2.5-14B actor models compared with GRPO. This observation can be attributed to the loss clip operation in AGPO, where only positive training loss from below-average actions is maintained while negative training loss is controlled in the gradient update. Consequently, the probability distribution is drifted upward asymmetrically by adequate gradient without being drifted downward meanwhile, which manifests higher measured entropy during training. The higher entropy eventually facilitates the generation of more diversified samples within the batch, which is essential for large-scale RL training.

\begin{figure}[htbp]
    \centering
    \includegraphics[width=0.5\textwidth]{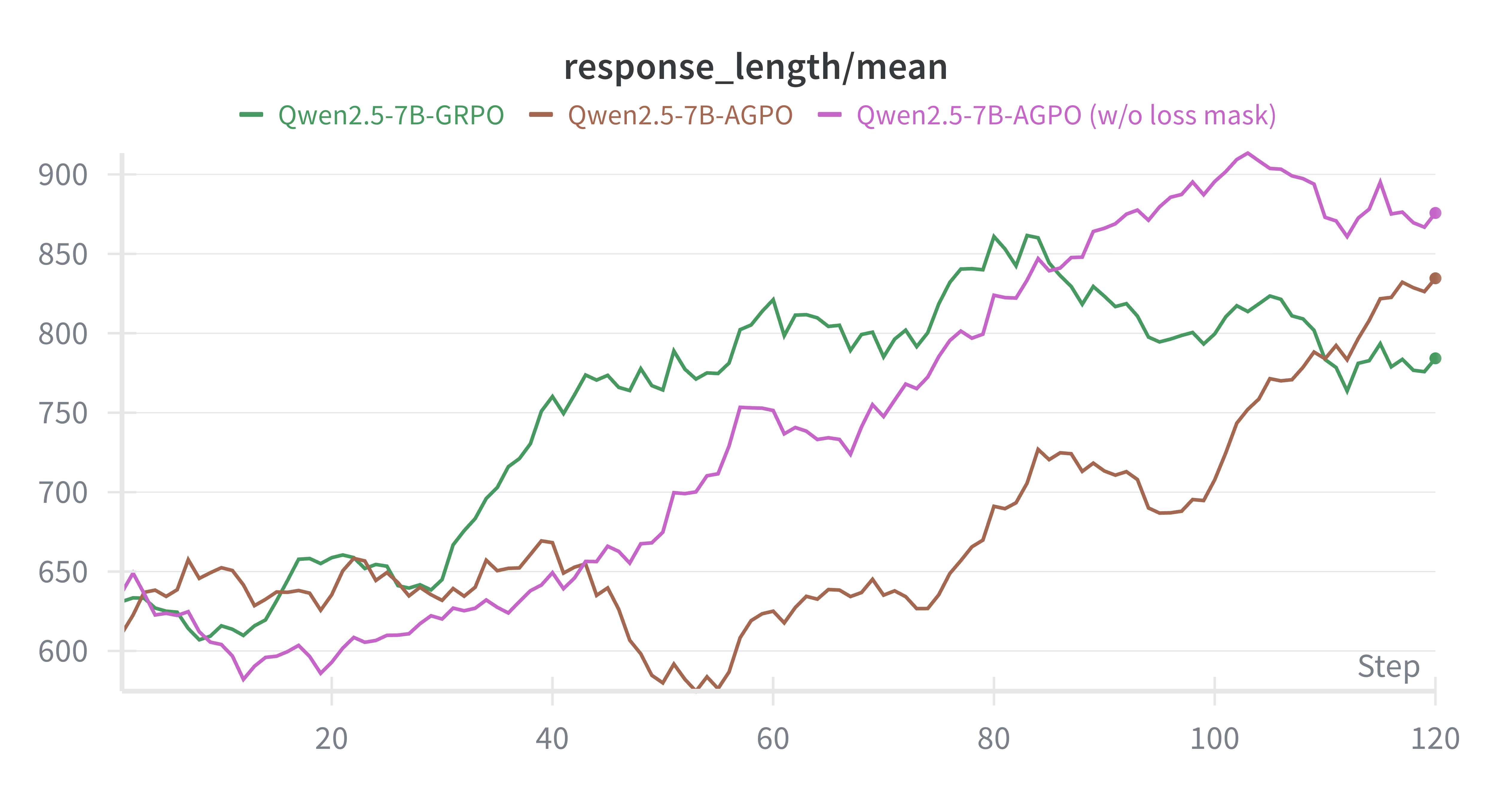}
    \caption{Response length curves of GRPO and AGPO for Qwen2.5-7B}
    \label{fig:7B-response_length}
\end{figure}

\begin{figure}[htbp]
    \centering
    \includegraphics[width=0.5\textwidth]{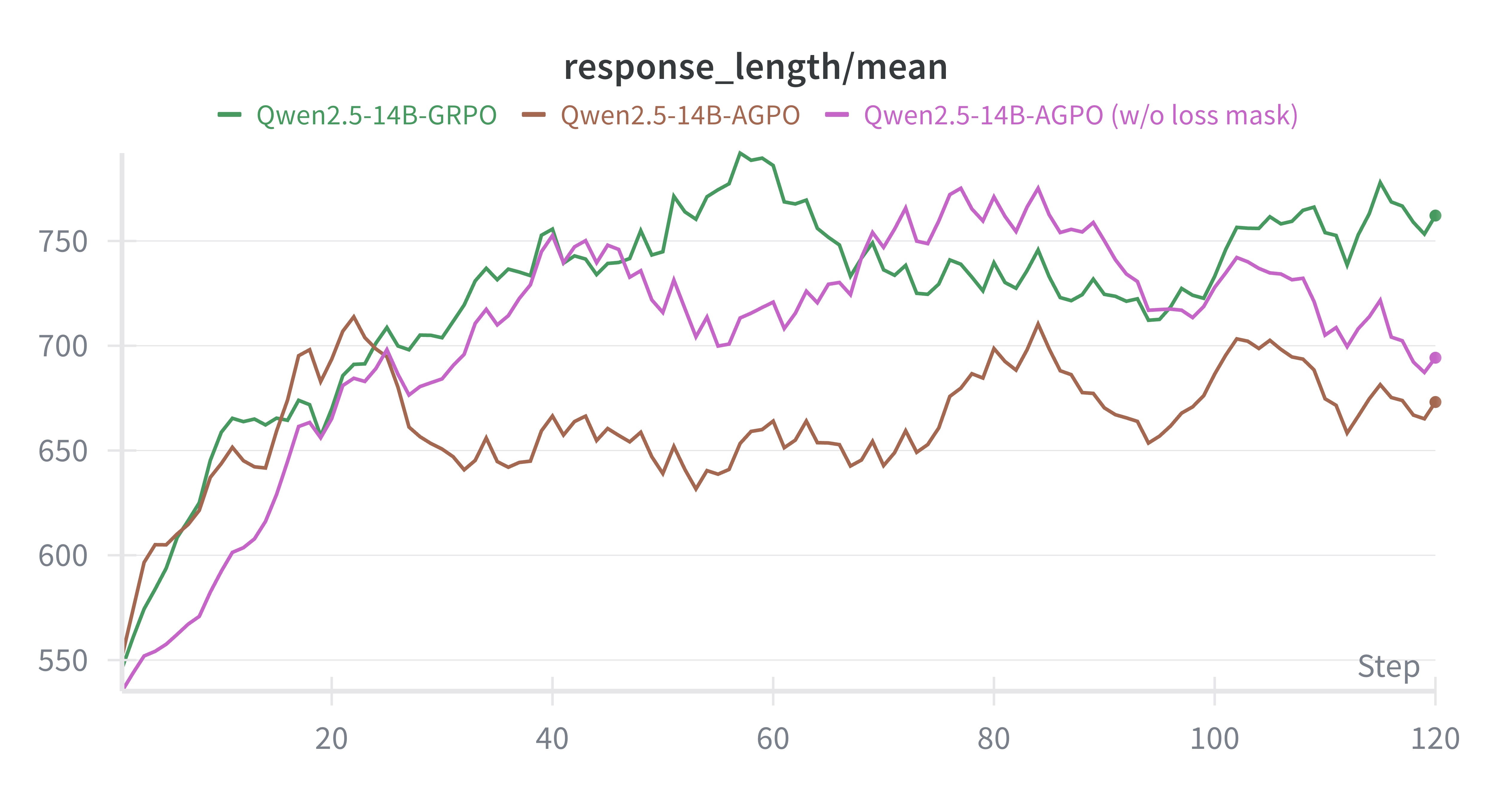}
    \caption{Response length curves of GRPO and AGPO for Qwen2.5-14B}
    \label{fig:14B-response_length}
\end{figure}

\begin{figure}[htbp]
    \centering
    \includegraphics[width=0.5\textwidth]{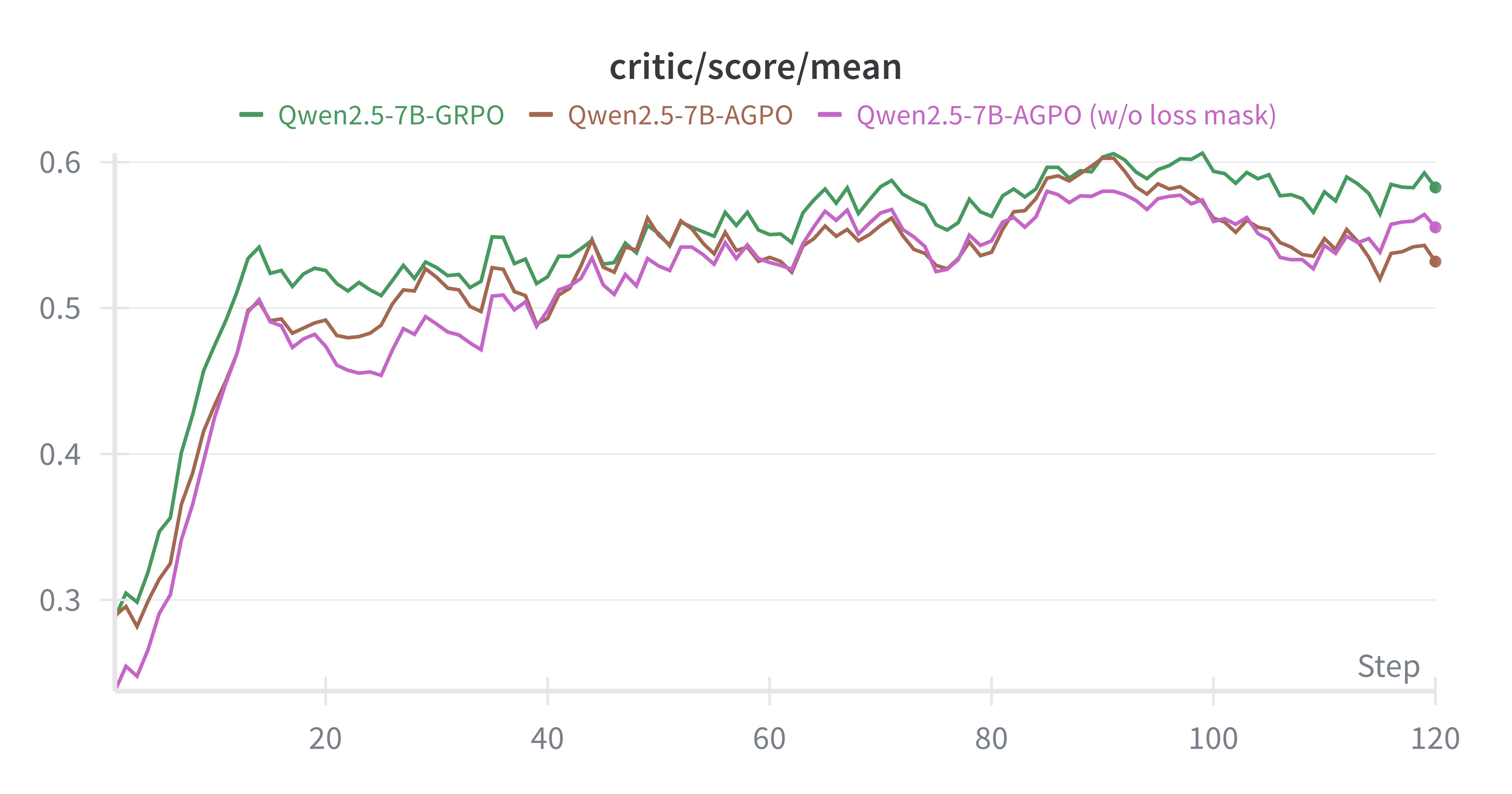}
    \caption{Reward score curves of GRPO and AGPO for Qwen2.5-7B}
    \label{fig:7B-critic_score}
\end{figure}

\begin{figure}[htbp]
    \centering
    \includegraphics[width=0.5\textwidth]{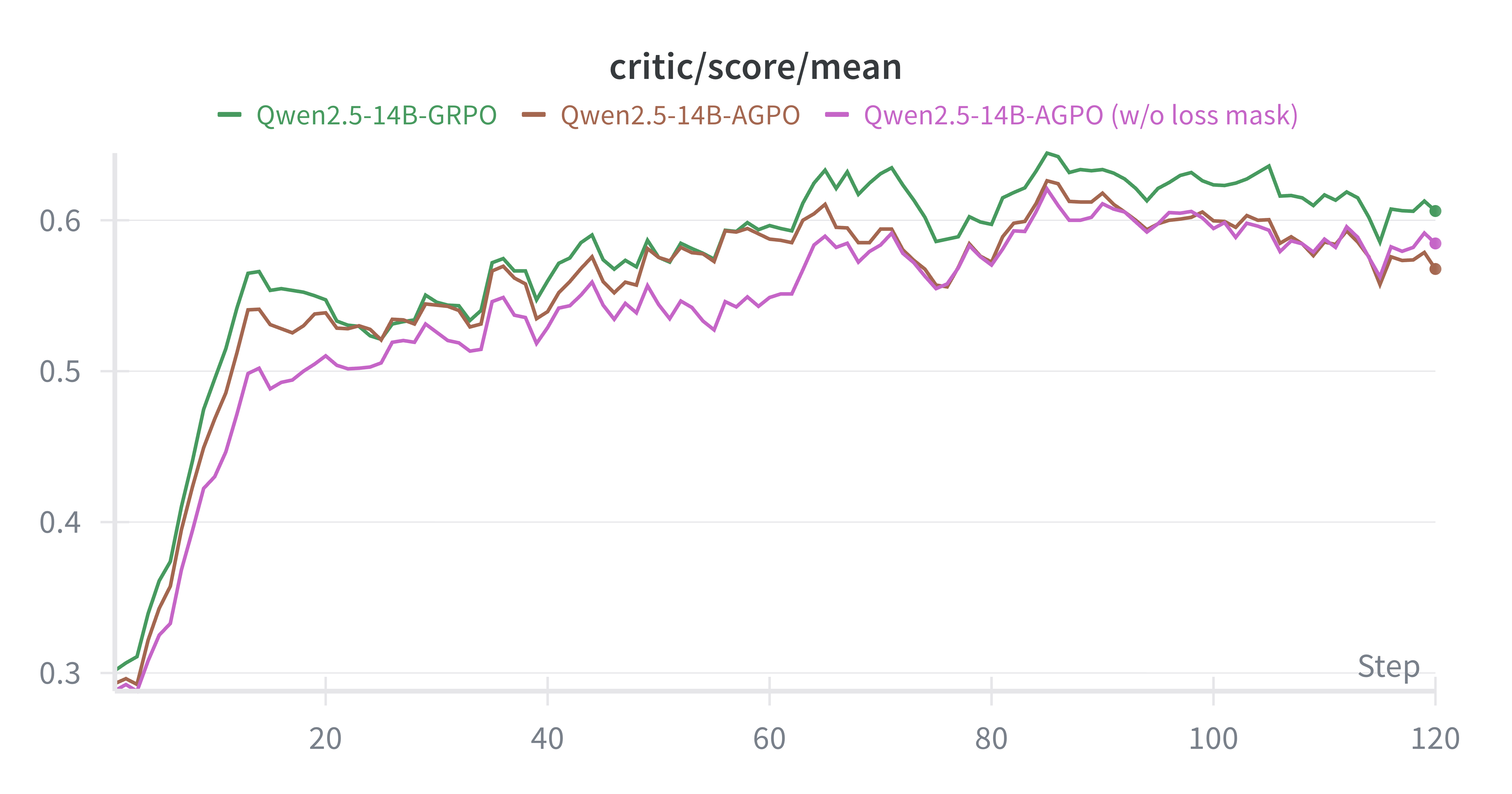}
    \caption{Reward score curves of GRPO and AGPO for Qwen2.5-14B}
    \label{fig:14B-critic_score}
\end{figure}

\autoref{fig:7B-response_length} and \autoref{fig:14B-response_length} indicate that substantial reductions in response length are achieved for both AGPO models. Meanwhile, comparable accuracy performance is maintained on training set as shown in \autoref{fig:7B-critic_score} and \autoref{fig:14B-critic_score}. We believe the loss mask implicitly serves as length-based reward to constrain response length, since the response length of AGPO is significantly shorter than that of AGPO (w/o loss mask). We also observe from the figures that the response length for AGPO is similar to GRPO at the early training stage, but eventually becomes shorter at middle and final stages compared with the GRPO baseline. This is because the initial model parameters are similar for both AGPO and GRPO. However, by masking all correct and all wrong groups where advantages are zero, the loss distribution within the batch is re-normalized as training progresses. This amplifies the gradient updates for AGPO so it tends to encourage the token distribution of correct response with higher confidence. Hence, AGPO requires fewer tokens or thinking words to arrive at the correct response, whereas GRPO needs more training iterations and response tokens to approach similar results.

\subsection{Generalization}
We conduct a set of additional experiments on code reasoning for assessing generalization of our method. Deepseek-R1-Distill-Qwen-7B is chosen as the base model for post RL training. We also curate 8.4K training data from CodeForce \citep{penedo2025codeforces} and CodeContest \citep{li2022competition}. We use LiveCodeBench (2024.10-2025.1) \citep{jain2024livecodebench} as the benchmark for evaluation. The results can be seen in \autoref{tab:code} that our method significantly outperforms GRPO by 3 points. We also observe the performance of GRPO drops dramatically from 40.1 to 23.5 after 40 steps, while the performance of AGPO keeps increasing.

\begin{table}[]
\resizebox{\columnwidth}{!}{
\begin{tabular}{lc}
\hline
\textbf{Model}                        & \textbf{LiveCodeBench (Pass@1)}   \\ \hline
Deepseek-R1-Distill-Qwen-7B           & 34.3                              \\
Deepseek-R1-Distill-Qwen-7B-GRPO	  & 40.1                              \\
Deepseek-R1-Distill-Qwen-7B-AGPO      & 43.1                              \\ \hline
\end{tabular}
}
\caption{Performance of different RL techniques on LiveCodeBench.}
\label{tab:code}
\end{table}

\section{Conclusion}
In this work, we propose a novel method, AGPO, to train a more powerful reasoning model. Our adaptive loss, including loss mask and clip, demonstrates noticeable improvement on both model performance and token efficiency. Our method also helps to avoid entropy collapse while training. As for future directions, we want to do more designs about the adaptive loss. For example, the current loss can be normalized to non-negative values by exponential equations.

\section*{Limitations}
We will experiment on more kinds of base models and datasets in future to validate universality of our method. More ablation studies around modifications will be taken as well. It is also uncertain if our approach can produce effects together with other tricks proposed by different GRPO refinements.

\newpage
\bibliography{acl}

\end{document}